\documentclass[11pt]{article}
\usepackage{acl2014}
\usepackage{times}
\usepackage{url}
\usepackage{latexsym}
\usepackage{amsmath}
\usepackage{multirow}
\usepackage{url}
\usepackage{array}
\usepackage{graphicx}
\usepackage[T5,T1]{fontenc}
\usepackage{algorithmic, amssymb, amsmath}   
\usepackage{algorithm}
\usepackage{multirow}
\usepackage{csquotes}
\usepackage{comment}

\title{Creating Lexical Resources for Endangered Languages}

\author{Khang Nhut Lam, Feras Al Tarouti and Jugal Kalita \\
Computer Science department \\
University of Colorado \\
1420 Austin Bluffs Pkwy, Colorado Springs, CO 80918, USA \\
{\tt \{klam2,faltarou,jkalita\}@uccs.edu}\\} %\\\And

\begin{document}
\maketitle
\begin{abstract}
This paper examines approaches to generate lexical resources for endangered languages. Our algorithms construct bilingual dictionaries and multilingual thesauruses using public Wordnets and a machine translator (MT). Since our work relies on only one bilingual dictionary between an endangered language and an ``intermediate helper'' language, it is applicable to languages that lack many existing resources.    
\end{abstract}

\section{Introduction}

Languages around the world are becoming extinct at a record rate.  The Ethnologue organization\footnote{http://www.ethnologue.com/} reports 424 languages as nearly extinct and 203 languages as dormant, out a total of 7,106 recorded languages. Many other languages are becoming endangered, a state which is likely to lead to their extinction, without determined intervention. According to UNESCO, \textquote{a language is endangered when its speakers cease to use it, use it in fewer and fewer domains, use fewer of its registers and speaking styles, and/or stop passing it on to the next generation...}. In America, UNESCO reports 134 endangered languages, e.g., Arapaho, Cherokee, Cheyenne, Potawatomi and Ute. 

One of the hallmarks of a living and thriving language is the existence and continued production of ``printed'' (now extended to online presence) resources such as books, magazines and educational materials in addition to oral traditions. There is some effort  afoot to document record and archive endangered languages. Documentation may involve creation of dictionaries, thesauruses, text and speech corpora. One possible way to resuscitate these languages is to make them more easily learnable for the younger generation. To learn languages and use them well, tools such as dictionaries and thesauruses are essential. Dictionaries are resources that empower the users and learners of a language. Dictionaries play a more substantial role than usual for endangered languages and are ``an instrument of language maintenance'' \cite{Gippert2006}. Thesauruses are resources that group words according to similarity \cite{Kilgarriff2003}. For speakers and students of an endangered language, multilingual thesauruses are also likely to be very helpful.

This study focuses on examining techniques that leverage existing resources for ``resource-rich'' languages to build lexical resources for low-resource languages, especially endangered languages. The only resource we need is a single available bilingual dictionary translating the given endangered language to English. First, we create a reverse dictionary from the input dictionary using the approach in \cite{Lam2013}. Then, we generate additional bilingual dictionaries translating from the given endangered language to several additional languages. Finally, we discuss the first steps to constructing multilingual thesauruses encompassing endangered and resources-rich languages.  To handle the word sense ambiguity problems, we exploit Wordnets in several languages. We experiment with two endangered languages: Cherokee and Cheyenne, and some resource-rich languages such as English, Finnish, French and Japanese\footnote{ISO 693-3 codes for Cherokee, Cheyenne, English, Finnish, French and Japanese are \emph{chr}, \emph{chy},  \emph{eng}, \emph{fin}, \emph{fra} and \emph{jpn}, respectively.}. Cherokee is the Iroquoian language spoken by 16,000 Cherokee people in Oklahoma and North Carolina. Cheyenne is a Native American language spoken by 2,100 Cheyenne people in Montana and Oklahoma.

The remainder of this paper is organized as follows. Dictionaries and thesauruses are introduced in Section 2. Section 3 discusses related work. In Section 4 and Section 5, we present approaches for creating new bilingual dictionaries and multilingual thesauruses, respectively. Experiments are described in Section 6. Section 7 concludes the paper.

\section{Dictionaries vs. Thesauruses}
A dictionary or a lexicon is a book (now, in electronic database formats as well) that consists of a list of entries sorted by the lexical unit. A lexical unit is a word or phrase being defined, also called \emph{definiendum}. A dictionary entry or a lexical entry simply contains a lexical unit and a definition \cite{Landau1984}. Given a lexical unit, the definition associated with it usually contains parts-of-speech (POS), pronunciations, meanings, example sentences showing the use of the source words and possibly additional information. A monolingual dictionary contains only one language such as The Oxford English Dictionary\footnote{http://www.oed.com/} while a bilingual dictionary consists of two languages such as the English-Cheyenne dictionary\footnote{http://cdkc.edu/cheyennedictionary/index-english/index.htm}. A lexical entry in the bilingual dictionary contains a lexical unit in a source language and equivalent words or multiword expressions in the target language along with optional additional information. A bilingual dictionary may be unidirectional or bidirectional.

Thesauruses are specialized dictionaries that store synonyms and antonyms of selected words in a language. Thus, a thesaurus is a resource that groups words according to similarity \cite{Kilgarriff2003}. However, a thesaurus is different from a dictionary. \cite{Roget1911} describes the organizes of words in a thesaurus as \textquote{... not in alphabetical order as they are in a dictionary, but according to the ideas which they express.... The idea being given, to find the word, or words, by which that idea may be most fitly and aptly expressed. For this purpose, the words and phrases of the language are here classed, not according to their sound or their orthography, but strictly according to their signification}. 
Particularly, a thesaurus contains a set of descriptors, an indexing language, a classification scheme or a system vocabulary \cite{Soergel1974}. A thesaurus also consists of relationships among descriptors. Each descriptor is a term, a notation or another string of symbols used to designate the concept. %p27, 30
Examples of thesauruses are Roget's international Thesaurus \cite{Roget2008}, the Open Thesaurus\footnote{http://www.openthesaurus.de/} or the one at thesaurus.com.

We believe that the lexical resources we create are likely to help endangered languages in several ways. These can be educational tools for language learning within and outside the community of speakers of the language. The dictionaries and thesauruses we create can be of help in developing parsers for these languages, in addition to assisting machine or human translators to translate rich oral or possibly limited written traditions of these languages into other languages. We may be also able to construct mini pocket dictionaries for travelers and students.

\section{Related work}
Previous approaches to create new bilingual dictionaries use intermediate dictionaries  to find chains of words with the same meaning. Then, several approaches are used to mitigate the effect of ambiguity. These include consulting the dictionary in the reverse direction \cite{Tanaka1994} and computing ranking scores, variously called a semantic score \cite{Bond2008},  an overlapping constraint score, a similarity score \cite{Paik2004} and a converse mapping score \cite{Shaw2013}. Other techniques to handle the ambiguity problem are merging results from several approaches: merging candidates from lexical triangulation \cite{Gollins2001}, creating a link structure among words \cite{Ahn2006} and building graphs connecting translations of words in several languages \cite{Mausam2010}. Researchers also merge information from several sources such as bilingual dictionaries and corpora  \cite{Otero2010} or a Wordnet \cite{Istvan2009} and \cite{Lam2013}.  Some researchers also extract bilingual dictionaries from corpora \cite{Ljubesic2011} and \cite{Bouamor2013}. The primary similarity among these methods is that either they work with languages that already possess several lexical resources or these approaches take advantage of related languages (that have some lexical resources) by using such languages as intermediary. The accuracies of bilingual dictionaries created from several available dictionaries and Wordnets are usually high. However, it is expensive to create such original lexical resources and they do not always exist for many languages. For instance, we cannot find any Wordnet for \emph{chr} or \emph{chy}. In addition, these existing approaches can only generate one or just a few new bilingual dictionaries from at least two existing bilingual dictionaries.

 \cite{Crouch1990} clusters documents first  using a complete link clustering algorithm and generates thesaurus classes or synonym lists based on user-supplied parameters such as a threshold  similarity value, number of documents in a cluster, minimum document frequency and specification of a class formation method. \cite{Curran2002a} and \cite{Curran2002b} evaluate performance and efficiency of thesaurus extraction methods and also propose an approximation method that provides for better time complexity with little loss in performance accuracy. \cite{Ramirez2013} develop a multilingual Japanese-English-Spanish thesaurus using freely available resources: Wikipedia and Wordnet. They extract translation tuples from Wikipedia from articles in these languages, disambiguate them by mapping to Wordnet senses, and extract a multilingual thesaurus with a total of 25,375 entries.

One thing to note about all these approaches is that they are resource hungry. For example, \cite{Lin1998} works with a 64-million word English corpus to produce a high quality thesaurus with about 10,000 entries. \cite{Ramirez2013} has the entire Wikipedia at their disposal with millions of articles in three languages, although for experiments they use only about 13,000 articles in total. When we work with endangered or low-resource languages, we do not have the luxury of collecting such  big corpora or accessing even a few thousand articles from Wikipedia or the entire Web. Many such languages have no or very limited Web presence. As a result, we have to work with whatever limited resources are available.

\section{Creating new bilingual dictionaries}
 
A dictionary \emph{Dict(S,T)} between a source language \emph{S} and a target language \emph{T} has a list of entries. Each entry contains a word \emph{s} in the source language \emph{S}, part-of-speech (POS) and one or more translations in the target language \emph{T}. We call such a translation \emph{t}.
Thus, a dictionary entry is of the form \emph{<$s_i$,POS,$t_{i1}$>}, \emph{<$s_i$,POS,$t_{i2}$>}, ....

This section examines approaches to create new bilingual dictionaries for endangered languages from just one dictionary \emph{Dict(S,I)}, where \emph{S} is the endangered source language and \emph{I} is an ``intermediate helper'' language. We require that the language \emph{I} has an available Wordnet linked to the Princeton Wordnet (PWN) \cite{Fellbaum1998}. Many endangered languages have a bilingual dictionary, usually to or from a resource-rich language like French or English which is the intermediate helper language in our experiments. We make an assumption that we can find only one unidirectional bilingual dictionary translating from a given endangered language to English.

\subsection{Generating a reverse bilingual dictionary}
Given a unidirectional dictionary \emph{Dict(S,I)} or \emph{Dict(I,S)}, we reverse the direction of the entries to produce \emph{Dict(I,S)} or \emph{Dict(S,I)}, respectively. We apply an approach called Direct Reversal with Similarity (DRwS), proposed in \cite{Lam2013} to create a reverse bilingual dictionary from an input dictionary. 

The DRwS approach computes the distance between translations of entries by measuring their semantic similarity, the so-called \emph{simValue}.  The \emph{simValue} between two phrases is calculated by comparing the similarity of the \emph{ExpansionSet} for every word in one phrase with \emph{ExpansionSet} of every word in the other phrase. An \emph{ExpansionSet} of a phrase is a union of the synset, synonym set, hyponym set, and/or hypernym set of every word in it. The synset, synonym, hyponym and hypernym sets of a word are obtained from PWN. The greater is the \emph{simValue} between two phrases, the more semantically similar are these phrases. According to \cite{Lam2013}, if the \emph{simValue} is equal to or greater than 0.9, the DRwS approach produces the ``best'' reverse dictionary. 

\begin{figure}[!h]
\centering
\includegraphics[width=0.5\textwidth]{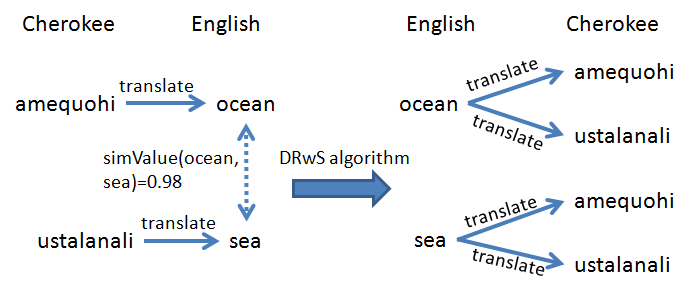} 
\caption{Example of creating entries for a reverse dictionary \emph{Dict(eng,chr)} from \emph{Dict(chr,eng)}. The \emph{simValue} between the words "ocean" and "sea" is 0.98, which is greater than the threshold of 0.90. Therefore, the words "ocean" and "sea" in English are hypothesized to have both meanings "amequohi" and "ustalanali" in Cherokee. We add these entries to  \emph{Dict(eng, chr)}. } 
\label{fig:ExampleDRwS }
\end{figure}

For creating a reverse dictionary, we skip entries with multiword expression in the translation. Based on our experiments, we have found that approach is successful and hence, it may be an effective way to automatically create a new bilingual dictionary from an existing one. Figure \ref{fig:ExampleDRwS } presents an example of generating entries for the reverse  dictionary.

\subsection{Building bilingual dictionaries to/from additional languages}
\begin{figure}[!h]
\centering
\includegraphics[width=0.5\textwidth]{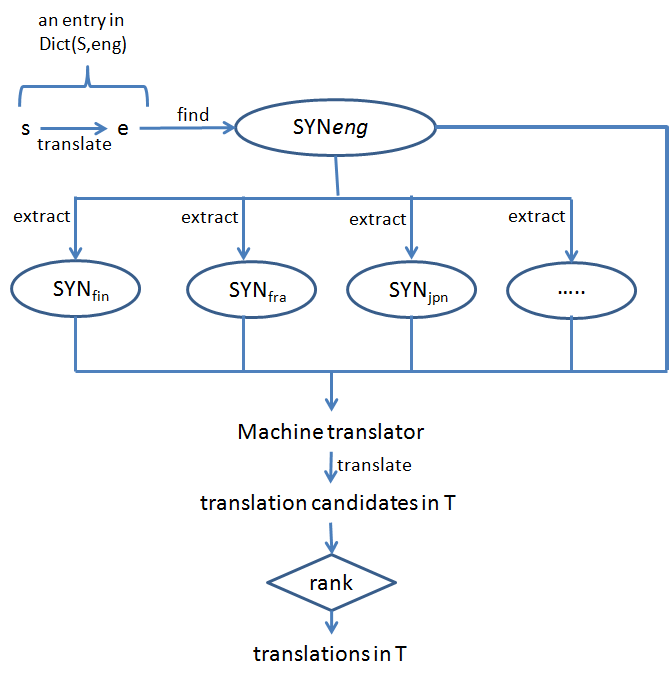} 
\caption{The approach for creating new bilingual dictionaries from intermediate Wordnets and a MT.}
\label{fig:createDics}
\end{figure}

We propose an approach using public Wordnets and MT to create new bilingual dictionaries \emph{Dict(S,T)} from an input dictionary \emph{Dict(S,I)}. As previously mentioned, \emph{I} is English in our experiments. \emph{Dict(S,T)} translates a word in an endangered language \emph{S} to a word or multiword expression in a target language \emph{T}.  In particular, we create bilingual dictionaries for an endangered language \emph{S} from a given dictionary \emph{Dict(S,eng)}. Figure \ref{fig:createDics} presents the approach to create new bilingual dictionaries.

\begin{figure}[!h]
\centering
\includegraphics[width=0.5\textwidth]{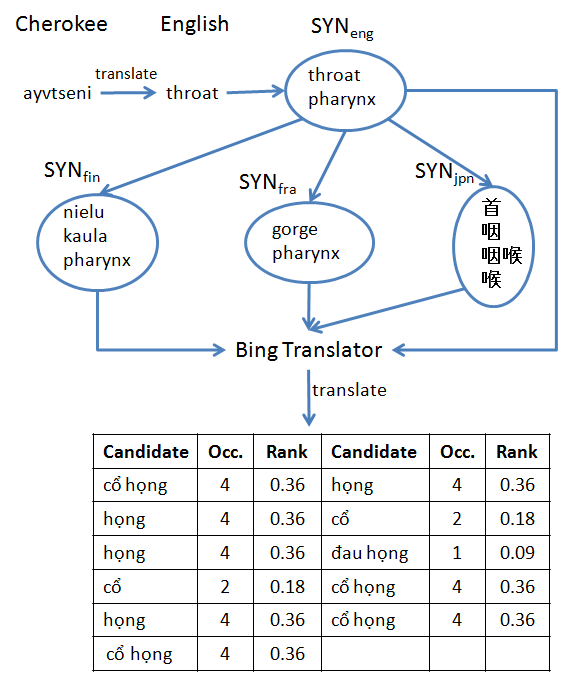}
\caption{Example of generating new entries for \emph{Dict(chr,vie)} from \emph{Dict(chr,eng)}. The word "ayvtseni" in \emph{chr} is translated to "throat" in \emph{eng}. We find all synonym words for "throat" in English to generate $SYN_{eng}$ and all synonyms in \emph{fin}, \emph{fra} and \emph{jpn} for all words in $SYN_{eng}$. Then, we translate all words in all $SYN_L$s to \emph{vie} and rank them. According to rank calculations, the best translations of  "ayvtseni" in \emph{chr} are the words "{\fontencoding{T5}\selectfont
c\h\ocircumflex} {\fontencoding{T5}\selectfont h\d{o}ng}"  and "{\fontencoding{T5}\selectfont h\d{o}ng}" in \emph{vie}.}
\label{fig:example}
\end{figure}

For each entry pair \emph{(s,e)} in a given dictionary \emph{Dict(S,eng)}, we find all synonym words of the word \emph{e} to create a list of synonym words in English: $SYN_{eng}$. $SYN_{eng}$ of the word \emph{eng} is obtained from the PWN. Then, we find all synonyms of words belonging to $SYN_{eng}$ in several non-English languages to generate $SYN_L$, $L\in\{fin, fra, jpn\}$. $SYN_{L}$ in the language \emph{L} is extracted from the publicly available Wordnet in language \emph{L} linked to the PWN. Next, translation candidates are generated by translating all words in $SYN_{L}$, $L \in\{\emph{eng, fin, fra, jpn}\}$ to the target language \emph{T} using an MT. A translation candidate is considered a correct translation of the source word in the target language if its rank is greater than a threshold. For each word \emph{s}, we may have many candidates. A translation candidate with a higher rank is more likely to become a correct translation in the target language. %Candidates having the same ranks are treated similarly. 
The rank of a candidate is computed by dividing its occurrence count by the total number of candidates. Figure \ref{fig:example} shows an example of creating entries for \emph{Dict(chr,vie)}, where \emph{vie} is Vietnamese, from \emph{Dict(chr,eng)}.

 \section{Constructing thesauruses}
As previously mentioned, we want to generate a multilingual thesaurus \emph{THS} composed of endangered and resource-rich languages. For example, we build the thesaurus encompassing an endangered language \emph{S} and \emph{eng}, \emph{fin}, \emph{fra} and \emph{jpn}. Our thesaurus contains a list of entries. Every entry has a unique \emph{ID}. Each entry is a 7-tuple: \emph{ID}, $SYN_{S}$, $SYN_{eng}$, $SYN_{fin}$, $SYN_{fra}$, $SYN_{jpn}$ and POS. Each $SYN_L$ contains words that have the same sense in language \emph{L}. All $SYN_L$, \emph{L} $\in$ \{\emph{S, eng, fin, fra, jpn}\} with the same ID have the same sense. 

\begin{figure}[!h]
\centering
\includegraphics[width=0.5\textwidth]{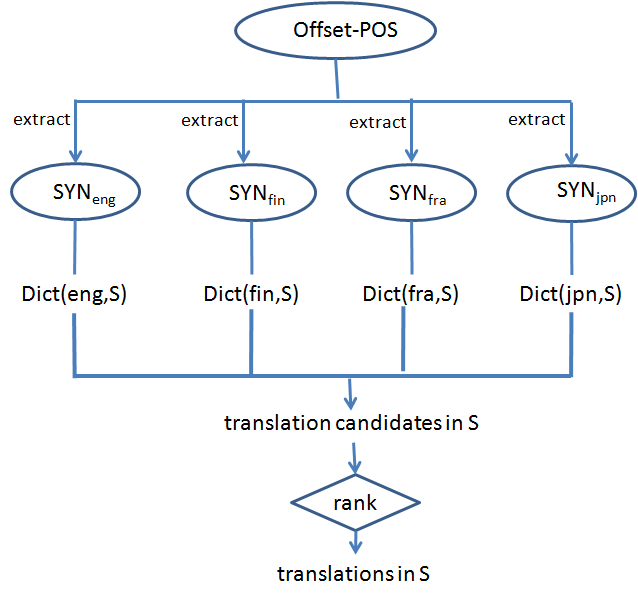}
\caption{The approach to construct a multilingual thesaurus encompassing an endangered language \emph{S} and resource-rich language.}
\label{fig:THSbaseline}
\end{figure}

This section presents the initial steps in constructing multilingual thesauruses using Wordnets and the bilingual dictionaries we create. The approach to create a multilingual thesaurus encompassing an endangered language and several resource-rich languages is presented in Figure \ref{fig:THSbaseline} and Algorithm \ref{alg:alg}.

First, we extract $SYN_L$ in resource-rich languages from Wordnets. To extract $SYN_{eng}$, $SYN_{fin}$, $SYN_{fra}$ and $SYN_{jpn}$, we use PWN and Wordnets linked to the PWN provided by the Open Multilingual Wordnet\footnote{http://compling.hss.ntu.edu.sg/omw/} project \cite{Bond2013}: FinnWordnet (FWN) \cite{Linden2010}, WOLF  (WWN) \cite{Sagot2008} and JapaneseWordnet (JWN) \cite{Isahara2008}. For each \emph{Offset-POS}, we extract its corresponding synsets from PWN, FWN, WWN and JWN to generate  $SYN_{eng}$, $SYN_{fin}$, $SYN_{fra}$ and $SYN_{jpn}$ (lines 7-10). The POS of the entry is the POS extracted from the \emph{Offset-POS} (line 5).  Since these Wordnets are aligned, a specific \emph{offset-POS} retrieves synsets that are equivalent sense-wise. Then, we translate all $SYN_L$s to the given endangered language \emph{S} using bilingual dictionaries we created in the previous section (lines 11-14). Finally, we rank translation candidates and add the correct translations to $SYN_S$ (lines 15-19). The rank of a candidate is computed by dividing its occurrence count by the total number of candidates. If a candidate has a rank value greater than a threshold, we accept it as a correct translation and add it to $SYN_S$.

\begin{algorithm}[!h]
\caption{}
Input: Endangered language S, PWN, FWN, WWN, JWN, Dict(eng,S), Dict(fin,S), Dict(fra,S) and Dict(jpn,S)\\
Output: thesaurus \emph{THS} 
\begin{algorithmic}[1]
\STATE ID:=0
\FORALL {\emph{offset-POS}s in PWN}
\STATE  ID++
\STATE candidates := $\phi$
\STATE $POS$=extract(offset-POS)
\STATE $SYN_{S}$:= $\phi$
\STATE $SYN_{eng}$=extract(offset-POS, PWN)
\STATE $SYN_{fin}$=extract(offset-POS, FWN)
\STATE $SYN_{fra}$=extract(offset-POS, WWN)
\STATE $SYN_{jpn}$=extract(offset-POS, JWN)
\STATE candidates+=translate($SYN_{eng}$,S)
\STATE candidates+=translate($SYN_{fin}$,S)
\STATE candidates+=translate($SYN_{fra}$,S)
\STATE candidates+=translate($SYN_{jpn}$,S)
\FORALL {candidate in candidates}
\IF{rank(candidate) > $\alpha$}
\STATE add(candidate,$SYN_S$)
\ENDIF
\ENDFOR
\STATE add ID, \emph{POS}  and all $SYN_L$ into \emph{THS}
\ENDFOR
 
\end{algorithmic}
 \label{alg:alg}
\end{algorithm}

\begin{figure}[!h]
\centering
\includegraphics[width=0.5\textwidth]{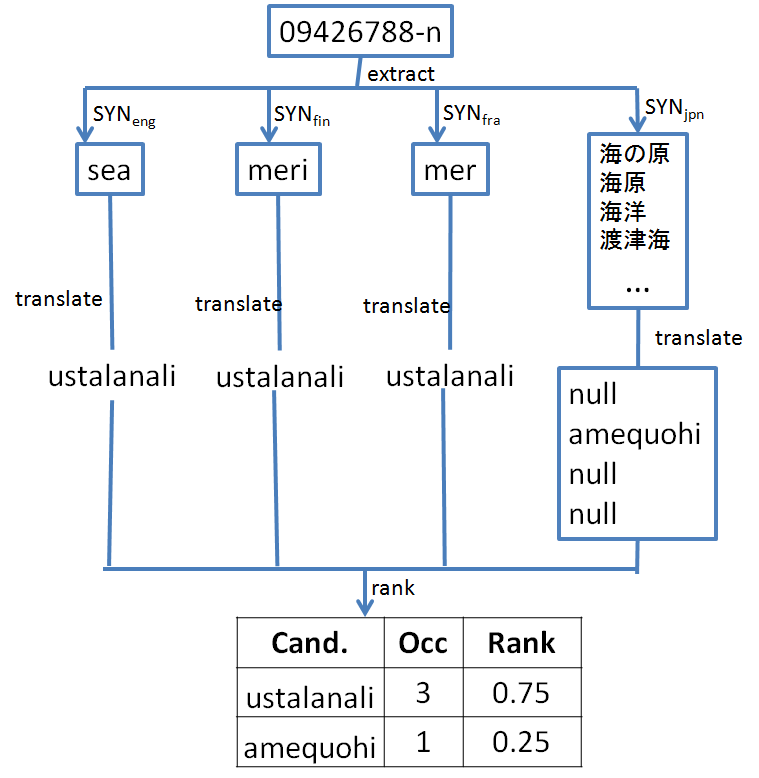}
\caption{Example of generating an entry in the multilingual thesaurus encompassing Cherokee, English, Finnish, French and Japanese.}
\label{fig:exampleTHS}
\end{figure}

Figure \ref{fig:exampleTHS} presents an example of creating an entry for the thesaurus.  We generate entries for the multilingual thesaurus encompassing of Cherokee, English, Finnish, French and Japanese. 

\begin{figure}[!h]
\centering
\includegraphics[width=0.5\textwidth]{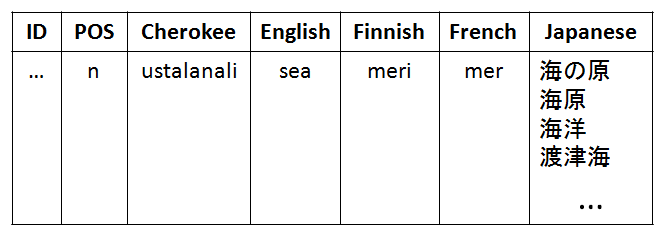}
\caption{An entry of the multilingual thesaurus encompassing Cherokee, English, Finnish, French and Japanese.}
\label{fig:resultTHS}
\end{figure}

We extract words belonging to \emph{offset-POS} "09426788-n" in PWN, FWN, WWN and JWN and add them into corresponding $SYN_L$. The POS of this entry is "n", which is a "noun". Next, we use the bilingual dictionaries we created to translate all words in $SYN_{eng}$, $SYN_{fin}$, $SYN_{fra}$, $SYN_{jpn}$ to the given endangered language, Cherokee, and rank them. According to the rank calculations, the best Cherokee translation is the word ``ustalanali''. The new entry added to the multilingual thesaurus is presented in Figure \ref{fig:resultTHS}.
  
\section{Experimental results} 

Ideally, evaluation should be performed by volunteers who are fluent in both source and destination languages. However, for evaluating created dictionaries and thesauruses, we could not recruit any individuals who are experts in two corresponding languages. We are in the process of finding volunteers who are fluent in both languages for some selected resources we create. 
\subsection{Datasets used}

We start with two bilingual dictionaries: \emph{Dict(chr,eng)}\footnote{http://www.manataka.org/page122.html} and \emph{Dict(chy,eng)}\footnote{http://www.cdkc.edu/cheyennedictionary/index-english/index.htm} that we obtain from Web pages. These are unidirectional bilingual dictionaries. The numbers of entries in \emph{Dict(chr,eng)} and \emph{Dict(chy,eng)} are 3,199 and 28,097, respectively. For entries in these input dictionaries without POS information, our algorithm chooses the best POS of the English word, which may lead to wrong translations. The Microsoft Translator Java API\footnote{https://datamarket.azure.com/dataset/bing/microsofttranslator} is used as another main resource. We were given free access to this API. We could not obtain free access to the API for the Google Translator.

The synonym lexicons are the synsets of PWN, FWN, JWN and WWN. Table \ref{tab:WordnetsInformation} provides some details of the Wordnets used. 

\begin{table}[!h]
 \centering
\begin{tabular}{|l|c|c|}
\hline    
\textbf{Wordnet }& \textbf{Synsets} & \textbf{Core}	\\
\hline    
JWN & 57,179 & 95\%\\ 
FWN &116,763 & 100\%\\
PWN& 117,659 & 100\%\\
WWN & 59,091 &92\%\\
\hline   
\end{tabular}
\caption{The number of synsets in the Wordnets linked to PWN 3.0 are obtained from the Open Multilingual Wordnet, along with the percentage of synsets covered from the semi-automatically compiled list of 5,000 "core" word senses in PWN. Note that synsets which are not linked to the PWN are not taken into account.}
\label{tab:WordnetsInformation}
\end{table}

\subsection{Creating reverse bilingual dictionaries}

From \emph{Dict(chr,eng)} and \emph{Dict(chy,eng)}, we create two reverse bilingual dictionaries \emph{Dict(eng,chr)} with 3,538 entries and \emph{Dict(eng,chy)} with 28,072 entries

\begin{figure}[!h]
\centering
\includegraphics[width=0.5\textwidth]{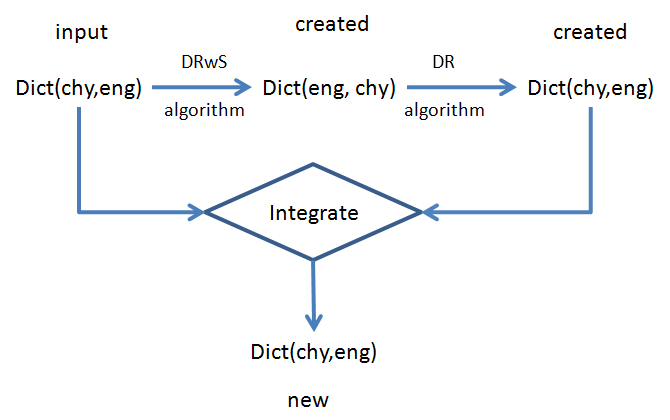}
\caption{Given a dictionary \emph{Dict(chy,eng)}, we create a new \emph{Dict(eng,chy)} using the DRwS approach of \cite{Lam2013}. Then, we create a new \emph{Dict(chy,eng)} using the DR approach from the created dictionary \emph{Dict(eng,chy)}. Finally, we integrate the generated dictionary \emph{Dict(chy,eng)} with the input dictionary \emph{Dict(chy,eng)} to create a new dictionary \emph{Dict(chy,eng)} with a greater number of entries}
\label{fig:reverseOfReverseDict}
\end{figure}

Next, we reverse the reverse dictionaries we produce to generate new reverse of the reverse (RR) dictionaries, then integrate the RR dictionaries with the input dictionaries to improve the sizes of dictionaries. During the process of generating new reverse dictionaries, we already computed the semantic similarity values among words to find words with the same meanings. We use a simple approach called the Direct Reversal (DR) approach in \cite{Lam2013} to create these RR dictionaries. To create a reverse dictionary \emph{Dict(T,S)}, the DR approach takes each entry \emph{<s,POS,t>} in the input dictionary \emph{Dict(S,T)} and simply swaps the positions of \emph{s} and  \emph{t}. The new entry \emph{<t,POS,s>} is added into \emph{Dict(T,S)}. Figure \ref{fig:reverseOfReverseDict} presents an example.

The number of entries in the integrated dictionaries \emph{Dict(chr,eng)} and \emph{Dict(chy,eng)} are 3,618 and 47,529, respectively. Thus, the number of entries in the original dictionaries have "magically" increased by 13.1\% and 69.21\%, respectively. 
 
\subsection{Creating additional bilingual dictionaries}
We can create dictionaries from \emph{chr} or \emph{chy} to any non-\emph{eng} language supported by the Microsoft Translator, e.g., Arabic (\emph{arb}), Chinese (\emph{cht}), Catalan (\emph{cat}), Danish (\emph{dan}), German (\emph{deu}), Hmong Daw (\emph{mww}), Indonesian (\emph{ind}), Malay (\emph{zlm}), Thai (\emph{tha}), Spanish (\emph{spa}) and \emph{vie}. Table \ref{tab:Dic3} presents the number of entries in the dictionaries we create. These dictionaries contain translations only with the highest ranks for each word.

\begin{table}[!h]
 \centering
\begin{tabular}{|l|r|l|r|}
\hline    

Dictionary&Entries&Dictionary&Entries\\ \hline
chr-arb&2,623&chr-cat&2,639\\
chr-cht&2,607&chr-dan&2,655\\
chr-deu&2,629&chr-mww&2,694\\
chr-ind&2,580&chr-zlm&2,633\\
chr-spa&2,607&chr-tha&2,645\\
chr-vie &2,618&chy-arb&10,604\\
chy-cat&10,748&chy-cht&10,538\\
chy-dan&10,654&chy-deu&10,708\\
chy-mww&10,790&chy-ind&10,434\\ %\hline
chy-zlm&10,690&chy-spa&10,580\\
chy-tha&10,696&chy-vie &10,848\\ \hline

\end{tabular}
\caption{The number of entries in some dictionaries we create.}
\label{tab:Dic3}
\end{table}
 
Although we have not evaluated entries in the particular dictionaries in Table 1, evaluation of dictionaries with non-endangered languages, but using the same approach, we have confidence that these dictionaries are of acceptable, if not very good quality.
 
\subsection{Creating multilingual thesauruses}
We construct two multilingual thesauruses: $THS_1$(\emph{chr, eng, fin, fra, jpn}) and $THS_2$(\emph{chy, eng, fin, fra, jpn}). The number of  entries in $THS_1$ and $THS_2$ are  5,073 and 10,046, respectively. These thesauruses we construct contain words with rank values above the average. A similar approach used to create Wordnet synsets \cite{Lam2014} has produced excellent results. We believe that our thesauruses reported in this paper are of acceptable quality. 

\subsection{How to evaluate}
Currently, we are not able to evaluate the dictionaries and thesauruses we create. In the future, we expect to evaluate our work using two methods. First, we will use the standard approach which is human evaluation to evaluate resources as previously mentioned.  Second, we will try to find an additional bilingual dictionary translating from an endangered language \emph{S} (viz., \emph{chr} or \emph{chy}) to another ``resource-rich'' non-English language (viz., \emph{fin} or \emph{fra}), then, create a new dictionary translating from \emph{S} to English using the approaches we have introduced. We plan to evaluate the new dictionary we create, say \emph{Dict(chr,eng)} against the existing dictionary \emph{Dict(chr,eng)}.

\section{Conclusion and future work} 
We examine approaches to create bilingual dictionaries and thesauruses for endangered languages from only one input dictionary, publicly available Wordnets and an MT. Taking advantage of available Wordnets linked to the PWN helps reduce ambiguities in dictionaries we create. We run experiments with two endangered languages: Cherokee and Cheyenne. We have also experimented with two additional endangered languages from Northeast India: Dimasa and Karbi, spoken by about 115,000 and 492,000 people, respectively. We believe that our research has the potential to increase the number of lexical resources for languages which do not have many existing resources to begin with. We are in the process of creating reverse dictionaries from bilingual dictionaries we have already created. We are also in the process of creating a Website where all dictionaries and thesauruses we create will be available, along with a user friendly interface to disseminate these resources to the wider public as well as to obtain feedback on individual entries. We will solicit feedback from communities that use the languages as mother-tongues. Our goal will be to use this feedback to improve the quality of the dictionaries and thesauruses. Some of resources we created can be downloaded from http://cs.uccs.edu/$\thicksim$linclab/projects.html


\begin{thebibliography}{}

\bibitem[\protect\citename{Kilgarriff}2003]{Kilgarriff2003}
Adam Kilgarriff.
\newblock 2003.
\newblock {Thesauruses for natural language processing}. 
\newblock In {\em Proceedings of the Joint Conference on Natural Language Processing and Knowledge Engineering}, pages 5--13, Beijing, China, October.

\bibitem[\protect\citename{Sagot and Fi\v{s}er}2008]{Sagot2008}
Benoit Sagot and Darja Fi\v{s}er.
\newblock 2008.
\newblock {Building a free {F}rench {W}ordnet from multilingual resources}.
\newblock In {\em Proceedings of OntoLex}, Marrakech, Morocco.


\bibitem[\protect\citename{Crouch}1990]{Crouch1990}
Carolyn J. Crouch
\newblock 1990.
\newblock {An approach to the automatic construction of global thesauri}, 
\newblock {\em Information Processing \& Management}, 26(5): 629--640.


\bibitem[\protect\citename{Fellbaum}1998]{Fellbaum1998}
Christiane Fellbaum.
\newblock 1998.
\newblock {\em Wordnet: An Electronic Lexical Database}.
\newblock  MIT Press, Cambridge, Massachusetts, USA.


\bibitem[\protect\citename{Soergel}1974]{Soergel1974}
Dagobert Soergel.
\newblock 1974.
\newblock {\em Indexing languages and thesauri: construction and maintenance}.
\newblock Melville Publishing Company, Los Angeles, California.

\bibitem[\protect\citename{Bouamor et al.}2013]{Bouamor2013}
Dhouha Bouamor, Nasredine Semmar and Pierre Zweigenbaum.
\newblock 2013
\newblock {Using Wordnet and Semantic Similarity for Bilingual Terminology Mining from Comparable Corpora}. 
\newblock In {\em Proceedings of the 6th Workshop on Building and Using Comparable Corpora}, pages 16--23, Sofia, Bulgaria, August. Association for Computational Linguistics.


\bibitem[\protect\citename{Lin}1998]{Lin1998}
Dekang Lin.
\newblock 1998.
\newblock {Automatic retrieval and clustering of similar words}. 
\newblock In {\em Proceedings of the 17th International Conference on Computational Linguistics (Volume 2)}, pages 768--774, Montreal, Quebec, Canada.

 

\bibitem[\protect\citename{Bond and Ogura}2008]{Bond2008}
Francis Bond and Kentaro Ogura.
\newblock 2008
\newblock {Combining linguistic resources to create a machine-tractable Japanese-Malay dictionary}. 
\newblock {\em Language Resources and Evaluation}, 42(2): 127--136.

 

\bibitem[\protect\citename{Bond and Foster}2013]{Bond2013}
 Francis  Bond and Ryan Foster.
\newblock 2013.
\newblock {Linking and extending an open multilingual {W}ordnet.}
\newblock In {\em Proceedings of 51st Annual Meeting of the Association for Computational Linguistics (ACL 2013)}, pages 1352--1362, Sofia, Bulgaria, August.

 


\bibitem[\protect\citename{Isahara \bgroup et al.\egroup }2008]{Isahara2008}
Hitoshi Isahara, Francis Bond, Kiyotaka Uchimoto, Masao Utiyama and Kyoko Kanzaki.
\newblock 2008.
\newblock {Development of {J}apanese {W}ordnet}.
\newblock In {\em Proceedings of 6th International Conference on Language Resources and Evaluation (LREC 2008)}, pages 2420--2423, Marrakech, Moroco, May.


\bibitem[\protect\citename{Curran and Moens}2002a]{Curran2002a}
James R. Curran and Marc Moens.
\newblock 2002a.
\newblock {Scaling context space}.
\newblock In {\em Proceedings of the 40th Annual Meeting of Association for Computational Linguistics (ACL 2002)}, pages 231--238, Philadelphia, USA, July.




\bibitem[\protect\citename{Curran and Moens}2002b]{Curran2002b}
James R. Curran and Marc Moens.
\newblock 2002b.
\newblock {Improvements in automatic thesaurus extraction},   
\newblock In {\em Proceedings of the Workshop on Unsupervised lexical acquisition (Volume 9)}, pages 59--66, Philadelphia, USA, July. Association for Computational Linguistics.



\bibitem[\protect\citename{Ram{\'\i}rez \bgroup et al.\egroup}2013]{Ramirez2013}
Jessica Ram{\'\i}rez, Masayuki Asahara and Yuji Matsumoto.
\newblock 2013.
\newblock {Japanese-Spanish thesaurus construction using English as a pivot}.
\newblock {\em arXiv preprint arXiv}:1303.1232.



\bibitem[\protect\citename{Gippert \bgroup et al.\egroup}2006]{Gippert2006}
Jost Gippert, Nikolaus Himmelmann and Ulrike Mosel, eds.
\newblock 2006.
\newblock {\em Essentials of Lnguage Documentation.}
\newblock Vol. 178, Walter de Gruyter GmbH \& Co. KG, Berlin, Germany.

\bibitem[\protect\citename{Lam and Kalita}2013]{Lam2013}
Khang N. Lam and Jugal Kalita.
\newblock 2013.
\newblock {Creating reverse bilingual dictionaries}. 
\newblock In {\em Proceedings of the Conference of the North American Chapter of the Association for Computational Linguistics: Human Language Technologies  (NAACL-HLT)}, pages 524--528, Atlanta, USA, June.

\bibitem[\protect\citename{Lam \bgroup et al.\egroup}2014]{Lam2014}
Khang N. Lam, Feras A. Tarouti and Jugal Kalita.
\newblock 2014.
\newblock {Automatically constructing {W}ordnet synsets}. 
\newblock To {\em appear at the 52nd Annual Meeting of the Association for Computational Linguistics (ACL 2014)}, Baltimore, USA, June.

\bibitem[\protect\citename{Ahn and Frampton}2006]{Ahn2006}
Kisuh Ahn and Matthew Frampton.
\newblock 2006.
\newblock {Automatic generation of translation dictionaries using intermediary languages}. 
\newblock In {\em Proceedings of the International Workshop on Cross-Language Knowledge Induction}, pages 41--44, Trento, Italy, April. European Chapter of the Association for Computational Linguistics.





\bibitem[\protect\citename{Lind\'en}2010]{Linden2010}
Krister Lind\'en and Lauri Carlson
\newblock 2010.
\newblock {{F}inn{W}ordnet - {W}ordNet p\r{a}finska via \"{o}vers\"{a}ttning, {L}exico{N}ordica.}
\newblock {\em Nordic Journal of Lexicography (Volume 17)}, pages 119--140.

\bibitem[\protect\citename{Tanaka and Umemura}1994]{Tanaka1994}
Kumiko Tanaka  and Kyoji Umemura.
\newblock 1994.
\newblock {Construction of bilingual dictionary intermediated by a third language}. 
\newblock In {\em Proceedings of the 15th Conference on Computational linguistics  (COLING 1994), Volume 1}, pages 297--303, Kyoto, Japan, August. Association for Computational Linguistics.


\bibitem[\protect\citename{Paik  \bgroup et al.\egroup }2004]{Paik2004}
Kyonghee Paik, Satoshi Shirai and Hiromi Nakaiwa.
\newblock 2004.
\newblock {Automatic construction of a transfer dictionary considering directionality}. 
\newblock In {\em Proceedings of the Workshop on Multilingual Linguistic Resources}, pages 31--38, Geneva, Switzerland, August . Association for Computational Linguistics.

 
 

\bibitem[\protect\citename{Mausam \bgroup et al.\egroup }2010]{Mausam2010}
Mausam, Stephen Soderland, Oren Etzioni, Daniel S. Weld, Kobi Reiter, Michael Skinner, Marcus Sammer and Jeff Bilmes
\newblock 2010.
\newblock {Panlingual lexical translation via probabilistic inference}.
\newblock {\em Artificial Intelligence}, 174(2010): 619--637.


\bibitem[\protect\citename{Ljube\v si\'c and Fi\v ser}2011]{Ljubesic2011}
Nikola Ljube\v si\'c and Darja Fi\v ser.
\newblock 2011.
\newblock {Bootstrapping bilingual lexicons from comparable corpora for closely related languages}. 
\newblock In {\em Proceedings of the 14th International Conference on Text, Speech and Dialogue (TSD 2011)}, pages 91--98. Plze\v n, Czech Republic, September.

\bibitem[\protect\citename{Otero and Campos}2010]{Otero2010}
Pablo G. Otero and Jos\'e R.P. Campos.
\newblock 2010.
\newblock {Automatic generation of bilingual dictionaries using intermediate languages and comparable corpora}. 
\newblock In {\em Proceedings of the 11th International Conference on Computational Linguistic and Intelligent Text Processing (CICLing'10 )}, pages 473--483, Ia\k{s}i, Romania, March.


\bibitem[\protect\citename{Roget}1911]{Roget1911}
Peter M. Roget.
\newblock 1911.
\newblock {\em Roget's Thesaurus of English Words and Phrases...}.
\newblock Thomas Y. Crowell Company, New York, USA.


\bibitem[\protect\citename{Roget}2008]{Roget2008}
Peter M. Roget.
\newblock 2008.
\newblock {\em Roget's International Thesaurus}, 3rd Edition.
\newblock  Oxford \& IBH Publishing Company Pvt, New Delhi, India.

 



\bibitem[\protect\citename{Shaw \bgroup et al.\egroup }2013]{Shaw2013}
Ryan Shaw, Anindya Datta, Debra VanderMeer and Kaushik Datta.
\newblock 2013.
\newblock {Building a scalable database - Driven Reverse Dictionary}.
\newblock {\em IEEE Transactions on Knowledge and Data Engineering}, 25(3): 528--540.

 

\bibitem[\protect\citename{Landau}1984]{Landau1984}
Sidney I. Landau
\newblock 1984.
\newblock {\em Dictionaries: the art and craft of lexicography}.
\newblock Charles Scribner's Sons, New York, USA.


 
\bibitem[\protect\citename{Gollins and Sanderson}2001]{Gollins2001}
Tim Gollins and Mark Sanderson.
\newblock 2001.
\newblock {Improving cross language information retrieval with triangulated translation}. 
\newblock In {\em Proceedings of the 24th Annual International ACM SIGIR Conference on Research and Development in Information Retrieval}, pages 90--95, New Orleans, Louisiana, USA, September.


\bibitem[\protect\citename{Istv\'an and Shoichi}2009]{Istvan2009}
Varga Istv\'an and Yokoyama Shoichi.
\newblock 2009.
\newblock {Bilingual dictionary generation for low-resourced language pairs}. 
\newblock In {\em Proceedings of the 2009 Conference on Empirical Methods in Natural Language Processing (Volume 2)}, pages 862--870, Singapore, August. Association for Computational Linguistics.


 








 
 




 


 
 




 


 
\end{thebibliography}
\end{document}